\documentclass{article} 
\usepackage{iclr2022_conference,times}

\usepackage{amsmath,amsfonts,bm}

\def\eqref#1{equation~\ref{#1}}

\def\1{\bm{1}}

\DeclareMathAlphabet{\mathsfit}{\encodingdefault}{\sfdefault}{m}{sl}
\SetMathAlphabet{\mathsfit}{bold}{\encodingdefault}{\sfdefault}{bx}{n}

\usepackage{hyperref}
\usepackage{url}
\usepackage{caption}
\usepackage{makecell} 
\usepackage{graphicx}
\usepackage[labelformat=simple]{subcaption}

\usepackage{amsmath}
\usepackage[export]{adjustbox}
\usepackage{caption}
\usepackage{subcaption}
\usepackage{tabularx}
\usepackage{wrapfig}
\usepackage{dblfloatfix}
\usepackage{amsmath,amsfonts,amssymb}
\usepackage{bm}
\usepackage{caption}

\usepackage{booktabs}
\usepackage{algorithm}
\usepackage{algorithmic}
\usepackage{xcolor}
\newcommand{\blue}[1]{\textcolor{blue}{#1}}

\title{rQdia: Regularizing Q-Value Distributions With Image Augmentation}
\iclrfinalcopy

\author{Sam Lerman\\
University of Rochester\\
Rochester, NY\\
\texttt{slerman@ur.rochester.edu}\\
\And
Jing Bi\\
University of Rochester\\
Rochester, NY\\
\texttt{jing.bi@rochester.edu}
}

\iclrfinalcopy 
\begin{document}

\maketitle

\begin{abstract}
rQdia (pronounced “Arcadia”) regularizes Q-value distributions with augmented images in pixel-based deep reinforcement learning. 
With a simple auxiliary loss,
that equalizes these distributions via MSE, 
rQdia boosts DrQ and SAC 
on $9/12$ and $10/12$ tasks respectively in the MuJoCo Continuous Control Suite from pixels, and Data-Efficient Rainbow on $18/26$ Atari Arcade environments. Gains are measured in both sample efficiency and longer-term training. Moreover, the addition of rQdia finally propels model-free continuous control from pixels over the state encoding baseline. 
\end{abstract}

\section{Introduction}

Human perception is invariant to and remarkably robust against many perturbations, like discoloration, obfuscation, and low exposure. On the other hand, artificial neural networks do not intrinsically carry these invariance properties, though some invariances may be induced architecturally through inductive biases like convolution, kernel rotation, and dilation. In deep reinforcement learning (RL) from pixels, an agent is tasked to learn from raw pixels and must therefore learn to visually interpret a scene. Thus, recent approaches in deep RL have turned to the self-supervision and data augmentation techniques found in computer vision. Indeed, such contrastive learning auxiliary losses~\citep{curl} or data augmentation regularizers~\citep{drq} have afforded greater sample efficiency and final scores in both the DeepMind Continuous Control Suite~\citep{dmc} from pixels and Atari Arcade Learning Environments~\citep{ale}.


Nevertheless, pixel-based approaches continue to lag behind models that learn directly from state embeddings, not just in terms of sample efficiency, but also in final reward performance. For example, the recent DrQ~\citep{drq}, an image augmentation-based regularizer to SAC~\citep{sac}, reports falling $14.5\%$ short of its state embedding-based SAC counterpart on the Cheetah Walk task. Such discrepancies indicate that visual representations are not yet up to par with state embeddings, at least not for locomotive continuous control. State embeddings have many properties that facilitate generalization, such as location invariance, and to a degree, invariance between morphological relations. If a subset of the dimensions of a state embedding always indicates feet position, then the relation of ``one foot in front of the other'' will be represented by those dimensions invariant to the placement of the robot’s arms, head, or other body parts. Parametric visual encodings are not guaranteed to learn such invariances with respect to the robot's morphology.

This begs the question of what other signals are available in deep reinforcement learning which can guide visual representation learning towards more invariant encodings. We propose that Q-value distributions, the expected cumulative discounted reward across \textit{distributions} of actions for a given state, offer such a signal. For a discrete set of actions $a_0, …, a_n$, a state $s_t$, and a Q-function $Q$, a Q-value distribution is simply:

$$Q(s_t, a_0), …, Q(s_t, a_n).$$

\begin{figure}
\centering
\includegraphics[width=.8\linewidth]{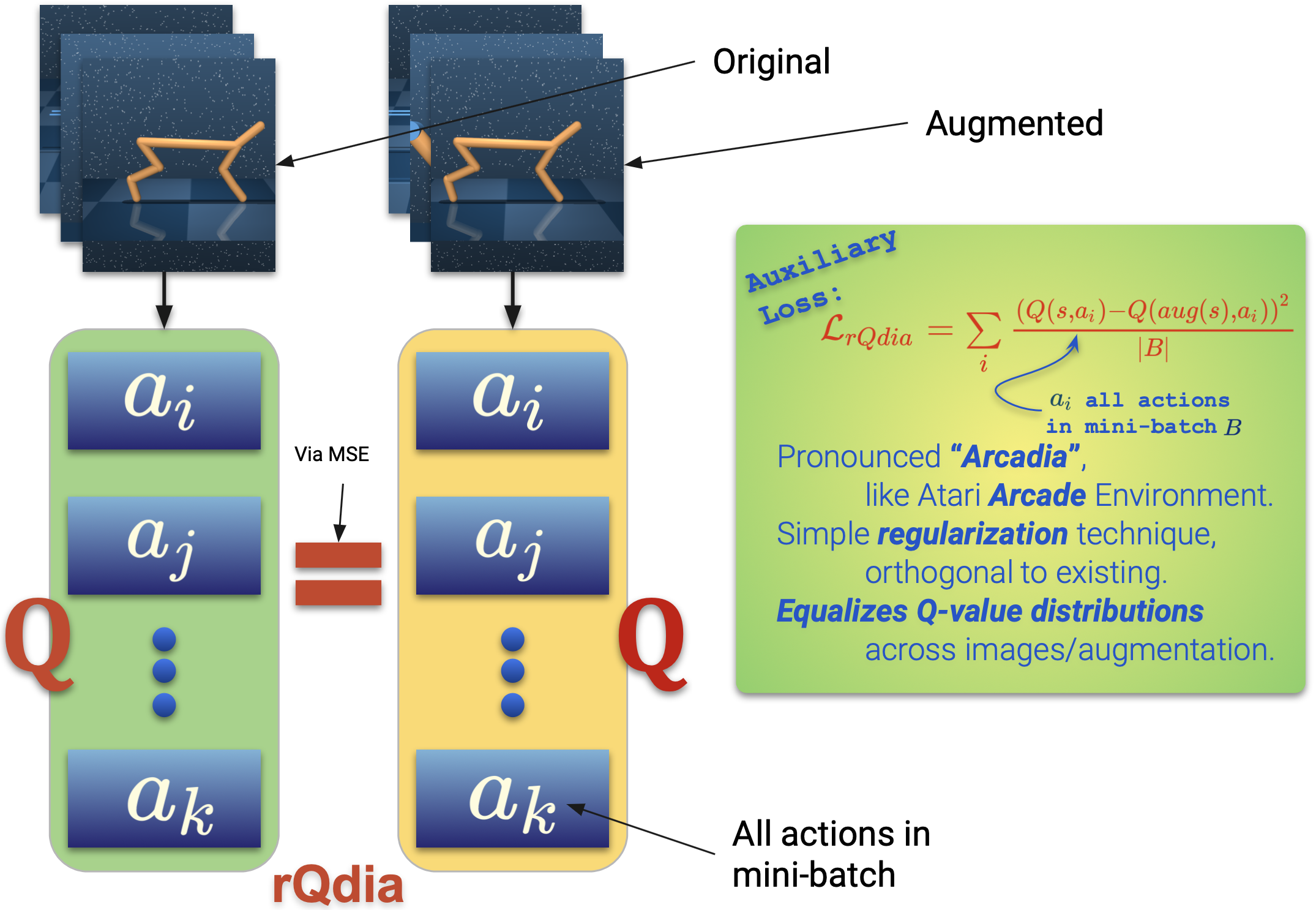}
\caption{\textit{\textbf{``rQdia in a nutshell"}} rQdia regularizes Q-value distributions across mini-batches of actions between images and their augmentations (\textit{e.g.}, \textit{crop}/\textit{translate}) with a simple auxiliary loss.
}
\label{rqdia}
\end{figure}

This distribution over actions for a state is a measure of the current \textit{and} future value of each action for that state. We review states, actions, rewards, Q-values, and the reinforcement learning Markov Decision Process formal framework in the subsequent Section \ref{MDP}. This distribution carries many of the same invariances afforded by state embeddings, if not some additional ones. For example, if the optimal action is to put ``one foot in front of the other,'' then the Q-value distribution reflects the preferability of this action over other actions regardless of (invariant to) where the agent is located, how it is holding its hands, or any other potentially superfluous contextual information. 

While recent work has found that individual Q-values benefit from regularizing across augmented images~\citep{drq}, we consider that Q-value \textit{distributions} are also important, as they contain information not just about one action in isolation, but multiple actions in relation to one another. 

\section{Background}

\subsection{Deep Reinforcement Learning From Pixels}
\label{MDP}


A Markov Decision Process (MDP) consists of an action $a \in {\rm I\!R}^{d_a}$, state $s \in {\rm I\!R}^{d_s}$, and reward $r \in {\rm I\!R}$. ``From pixels'' assumes that state $s$ is an image frame or multiple image frames stacked together. The action is sampled from a policy at any $t$ time step $a_t \sim \pi(s_t)$. Taking such actions yields a trajectory $\tau = (s_0, a_0, s_1, a_1, ..., s_T)$ via the dynamics model $s_{t+1} \sim f(s_t, a_t)$ of the environment and its rewards $r_{t+1} = \ R(s_t, a_t)$. The goal is to maximize cumulative discounted reward $R(\tau) = \sum_{t = 0}^T r_t \gamma^t$ where $\gamma$ is the temporal discount factor. The optimal action for a state $a^*(s) = argmax_a Q^*(s, a)$ thus depends on the state-action value function, also known as the Q-value, $Q^{\pi}(s, a) = \mathbf{E}_{\tau \sim \pi}[R(\tau) | s_0 = s, a_0 = a]$.

\subsection{Soft Actor-Critic}

Soft-Actor Critic (SAC)~\citep{sac} is an RL algorithm which learns a Q-value function $Q_\phi$ parameterized by $\phi$, optimized with one-step Bellman error, and a stochastic Gaussian policy $\pi_{\theta}$, optimized by maximizing $Q_\phi$ and entropy, where $\theta$ is the parameter subspace of $\phi$ corresponding to the policy network. In SAC, this Q-value function also includes entropy $H(s_t) = -\pi_{\theta}(s_t)log(\pi_{\theta}(s_t))$ as part of the reward. In practice, $Q_\phi$ and $\pi_{\theta}$ share a convolutional neural net encoder, and policy head $\pi_{\theta}$ is made differentiable via the reparameterization trick.

\subsection{Rainbow}

Rainbow~\citep{rainbow} similarly represents pixel-level inputs with a convolutional neural net, but operates on a discrete action space and thus directly estimates Q-values for each discrete action. To further improve performance, several refinements are used: these Q-values are sampled from a multivariate Gaussian probabilistically rather than deterministically~\citep{distributional}, double Q networks~\citep{ddqn}, noise is injected into the network parameters~\citep{noise}, dueling DQNs~\citep{duel}, n-step returns~\citep{nstep}, and mini-batches are sampled from a prioritized experience replay~\citep{per}.



\section{Related Work}

\subsection{Continuous Control From Images}

Continuous control from images is highly challenging as images are high-dimensional and supervisory signals in reinforcement learning (RL) are limited to scalar
rewards.
One way to mitigate this challenge is to learn the dynamics of the environment with model-based algorithms, then perform policy search through the learned model~\citep{model1, model2, model3}.  

More recently, model-free methods like SAC with a convolutional encoder received an enhancement by introducing an auxiliary image reconstruction task~\citep{modelfree3, curl, modelfree5, modelfree6}, which provides additional self-supervision for visual representation learning.


\subsection{Image Augmentation } 
Image augmentation is originally developed to counter over-fitting in computer vision, which applies transformations that include color shift, affinity translation, scale, etc.~\citep{cvia1,cvia2,cvia3,cvia4,cvia5}, to the input image where the output label is invariant.
In this way, the neural network's search space is restricted to converging to more invariant solutions during optimization, thus benefiting its generalizability. Such techniques have become largely standard in computer vision since AlexNet~\citep{alexnet}.
A similar idea is borrowed in RL for continuous control from images \citep{drq, curl} and discrete control Atari \cite{spr}.
However, the reward is not invariant to all transformations. For example, if the agent is rewarded for moving right, applying a rotation to the image may not preserve the reward. 
In~\citet{drq}, the authors investigated several common image transformations and conclude that random shifts strike a good balance between simplicity and performance.

\subsection{Contrastive Learning} 

Furthermore, contrastive learning aims at grouping similar samples closer and diverse samples farther from each other, which will yield more invariant visual representations that benefit recognition~\citep{cvia5, visioncl1, visioncl2, visioncl3}. 
Usually, this is done by adding auxiliary losses to classify  associated pairs created by image augmentation positively while discriminating random pairs negatively.
In the RL domain, CURL~\citep{curl} acquired higher performance on continuous control tasks from pixels by employing such an auxiliary loss in a SAC architecture.

\subsection{Regularization}

Many simple regularizers have benefited generalization in deep RL, including dropout~\citep{dropout}, L2~\citep{l2}, and most recently and commonly, entropy~\citep{entropy1, sac, entropy3, entropy4}. Very recently, image augmentation has been used to regularize Q-values in continuous control~\citep{drq}. However, these regularizations are fully complementary to rQdia. rQdia regularizes Q-value distributions rather than individual Q-values, and can be incorporated with a simple auxiliary loss rather than averaging Q-value predictions in the network inference itself.

Simple random translations have been found to be the most useful in the DMControl from pixels setting ~\citep{curl, drq}. A variety of different augmentations are useful for different games in the Procgen benchmark~\citep{augs}, and~\citet{drq} uses Intensity variation for the Atari environments.

\subsection{Generalization And Sample Efficiency}

Scene variation has been used to improve generalization in RL, such as in robotics~\citet{robotics}, Meta-World~\citep{meta}, and ProcGen~\citep{procgen}. Domain randomization ~\citep{dr1, dr2} similarly varies the scene except by modifying the dynamics of the environment simulator itself.

\section{Methods}

As mentioned previously, we use Q-value distributions to lend a reinforcement learner greater visual invariance. Now, recently DrQ used image augmentation to induce greater invariance. The augmentation used was a simple padding and crop, \textit{e.g.}, a translation. 

\begin{figure}
\centering
\includegraphics[width=.8\linewidth]{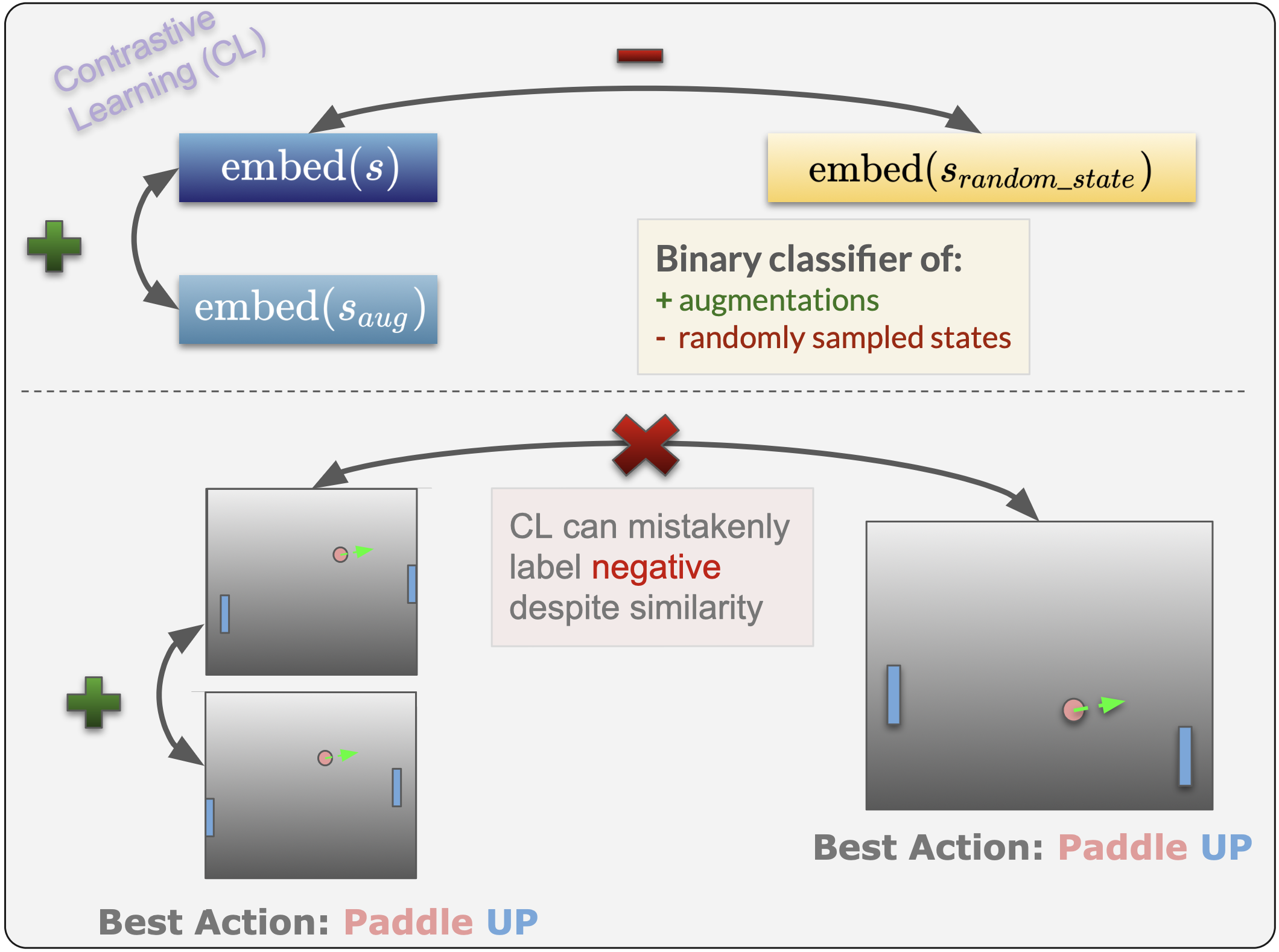}
\caption{Relying on contrastive learning has disadvantages that this emerging family of approaches based on simple equalization does not.}
\label{clproblem}
\end{figure}

Unsurprisingly, given what is known about data augmentation in computer vision, the addition of this small translation invariance helped visual representation learning. Interestingly, this benefit far exceeded that of the preceding contrastive learning based approach, where the augmented image was correlated with the current state, and randomly sampled historical images were ``contrasted.'' The latter, CURL~\citep{curl}, successfully adapted SAC~\citep{sac} to be compatible with pixel-level inputs instead of state embeddings. However, DrQ's direct use of the augmented image to predict Q-values outperformed the more conceptually complex contrastive learning approach. This is perhaps due to the greater freedom afforded to the model to map pixels to a latent space that corresponds directly with the end goal of evaluating the quality of an action at a given state, rather than inducing a decoupled latent space similarity with contrastive learning. Moreover, however, this direct use of the augmented image circumvents a major downside of naive contrastive learning, which is that negative samples are ``contrasted'' despite being sampled randomly from the agent's state history. This means that policy-relevant states may be de-associated despite the policy relevance. 
See Figure \ref{clproblem} for an example of this in the Atari environment Pong.

DrQ does not require an auxiliary loss, but rather just averages the Q-values of the anchor image with that of the augmented image, effectively training both in tandem. 

This is however optimized for a single Q-value, which is the one associated with the selected action. This is also the only Q-value for which we have a label, since we do not know the ground truth Q-value of an action that was not taken. 


Furthermore, we do not have a finite number of Q-values in environments such as the DeepMind Continuous Control Suite, where locomotive motor control is unrestricted spatially per joint and thus infinite.

Our solution in a nutshell: equalize Q-value distributions between anchors and augmentations by convolving all actions in the mini-batch. Why the mini-batch instead of randomly sampling actions? Simply because the mini-batch is already conveniently available during training and equates to using randomly sampled historical actions. By using historical actions, we can actually compute Q-value distributions with moderately realistic state-action pairs that could feasibly be encountered in simulation. We do not employ a heuristic for sampling such actions here in this work, but we note that sampling actions based on a more sophisticated measure, such as state similarity, is also possible.

Like DrQ, we use random translation as our augmentation technique. Specifically, we pad by 4 pixels, then crop randomly inward by 4 pixels. Without loss of generality, let's refer to this augmentation function as $aug(\cdot)$.

Further, let's define our state $s_t$ and batch size $n$. Then rQdia is simply:

\begin{equation}
\label{rqdiaeq}
\mathcal{L}_{rQdia} = \frac{1}{n} \sum\limits_{i < n} (Q(s_t, a_i) - Q(aug(s_t), a_i))^2,
\end{equation}

where $a_0, ..., a_{n-1}$ are the set of actions in the mini-batch.

Then this auxiliary loss is simply added to the RL agent's standard loss term. Voila, rQdia (summarized ``in a nutshell'' in Figure \ref{rqdia}). This is applied in parallel for each of the states in the mini-batch. If mini-batches are very large, it is possible to merely convolve a subset of the $n$ actions in the mini-batch for the $a_i$ in Equation \ref{rqdiaeq}.

Outside of continuous control, we just use the full discrete action space rather than convolving mini-batch-sampled actions. In Rainbow in particular, which uses distributional probabilistic Q-values, we also found that mean-squared error, as above in Equation \ref{rqdiaeq}, between Q-value logits, performed best, as opposed to using KL-divergence between their associated distributional probabilities, which would have been an alternative way to minimize the difference between the two Q-value distributions in Rainbow.




\begin{algorithm}[t] 
\caption{
rQdia added to original Soft Actor-Critic algorithm, pseudocode courtesy of~\citet{SpinningUp2018}. 
SAC is a good base framework and can be easily expanded to DrQ 
\citep{drq}.
} 
\label{rqdiacode} 
\begin{algorithmic}
\STATE Input: initial policy parameters $\theta$, Q-function parameters $\phi_1$, $\phi_2$, empty replay buffer $\mathcal{D}$ \STATE Set target parameters equal to main parameters
$\phi_{\text{targ},1} \leftarrow \phi_1$, $\phi_{\text{targ},2} \leftarrow \phi_2$ 
\STATE \blue{Denote augmentation function $aug(\cdot)$}
\REPEAT \STATE Observe state $s$ and select action $a \sim \pi_{\theta}(\cdot|s)$ \STATE Execute $a$ in the environment \STATE Observe next state $s'$, reward $r$, and done signal $d$ to indicate whether $s'$ is terminal \STATE Store $(s,a,r,s',d)$ in replay buffer $\mathcal{D}$ \STATE If $s'$ is terminal, reset environment state. \IF{it's time to update} \FOR{$j$ in range(however many updates)} \STATE Randomly sample a batch of transitions, $B = \{ (s,a,r,s',d) \}$ from $\mathcal{D}$ 
\STATE Compute targets for the Q functions: 
\begin{align*} y (r,s',d) &= r + \gamma (1-d) \left(\min_{i=1,2} Q_{\phi_{\text{targ}, i}} (s', \tilde{a}') - \alpha \log \pi_{\theta}(\tilde{a}'|s')\right), && \tilde{a}' \sim \pi_{\theta}(\cdot|s') 
\end{align*} 
\STATE Update Q-functions by one step of gradient descent using:
\begin{align*} & \nabla_{\phi_i} \frac{1}{|B|}\sum_{(s,a,r,s',d) \in B} \left( Q_{\phi_i}(s,a) - y(r,s',d) \right)^2 && \text{for } i=1,2 
\end{align*} 
\color{blue}
\STATE Update rQdia by one step of gradient descent using:
\begin{align*} 
& \nabla_{\phi_i} \frac{1}{|B|}\sum_{(s,a,...), (\hat{s}, \hat{a},...) \in B} \left( \min_{i=1,2}Q_{\phi_i}(s,\hat{a}) - \min_{i=1,2}Q_{\phi_i}(aug(s),\hat{a}) \right)^2 
\end{align*} 
\color{black}
\STATE Update policy by one step of gradient ascent using: \begin{equation*} \nabla_{\theta} \frac{1}{|B|}\sum_{s \in B} \Big(\min_{i=1,2} Q_{\phi_i}(s, \tilde{a}_{\theta}(s)) - \alpha \log \pi_{\theta} \left(\left. \tilde{a}_{\theta}(s) \right| s\right) \Big), \end{equation*} where $\tilde{a}_{\theta}(s)$ is a sample from $\pi_{\theta}(\cdot|s)$ which is differentiable wrt $\theta$ via the reparametrization trick. 
\STATE Update target networks with: 
\begin{align*} \phi_{\text{targ},i} &\leftarrow \rho \phi_{\text{targ}, i} + (1-\rho) \phi_i && \text{for } i=1,2 
\end{align*} 
\ENDFOR 
\ENDIF
\UNTIL{convergence} 
\STATE
\end{algorithmic} 
\end{algorithm}

Pseudo-code for adding rQdia to continuous control models like SAC-AE and DrQ is shown in Algorithm \ref{rqdiacode}. All code for continuous \textit{and} discrete action spaces will be released.

\begin{wrapfigure}{r}{0.48\textwidth}
\vspace{-2.7em}
  \begin{center}
    \includegraphics[width=0.48\textwidth]{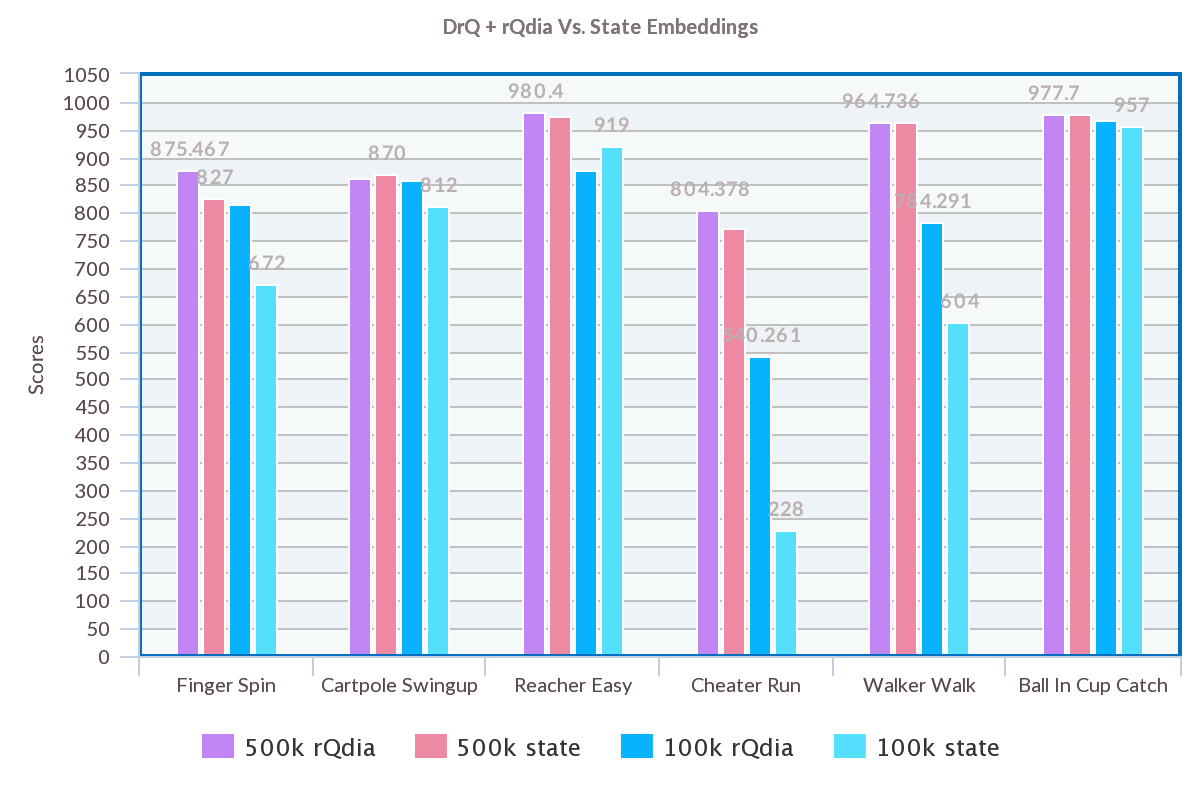}
  \end{center}
  \caption{DrQ + rQdia pushes DrQ over the top of the state embedding baseline.}
  \label{statechart}
  \vspace{-2em}
\end{wrapfigure}

\section{Experiments}
\label{experiments}

\subsection{100k Continuous Control}

rQdia added to DrQ yields several noteworthy results. Firstly, we find sample efficiency greatly improves with the addition of rQdia. In the 100k DeepMind Control Suite setting, rQdia benefits or improves over recent algorithms on 5/6 environments, as shown in Table \ref{rqdiachart}.

Interestingly, rQdia hurts performance on the finger spin environment when combined with DrQ, though improves over it when combined with SAC-AE as we will see later. We are not sure why this regularization cuts performance here, while it benefits performance in the more complex tasks or when combined with the older SAC-AE. 
Perhaps there is a tradeoff between generality and in-domain performance which would be observed if the training and testing settings were different or as difficulty increases. This could occur if the model overfits to an environment and relies on disagreeing/biased Q-value distributions.


Amazingly, in this 100k setting, rQdia substantially outperformed SAC from state embeddings. That is to say, that training from pixels actually learns \textit{faster} than training from state embeddings with rQdia. This is less often the case with other models (5/6 for DrQ + rQdia, 2/6 for DrQ with 1 approximately the same). We speculate that this is because rQdia induces original invariances early on in training that state embeddings do not possess, thanks to the added Q-value distribution invariance across image translatons. DrQ + rQdia is compared to state embeddings in Figure \ref{statechart}

\subsection{500k Continuous Control}

\begin{figure*}[t!]
    \centering
    \begin{subfigure}[t]{0.4\textwidth}
        \centering
        \includegraphics[height=1.6in]{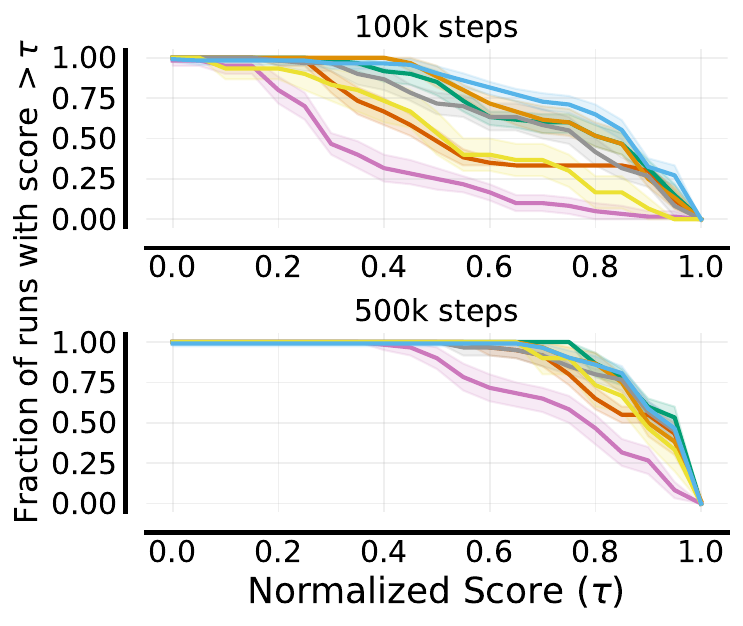}
        \caption{Lorem ipsum}
    \end{subfigure}%
    ~ 
    \begin{subfigure}[t]{0.5\textwidth}
        \centering
        \includegraphics[height=1.2in]{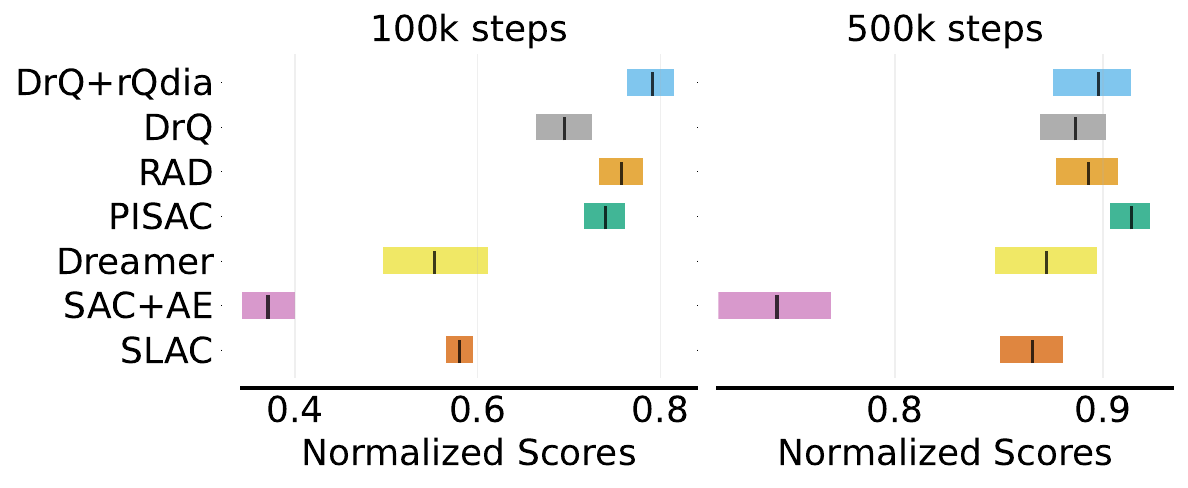}
        \caption{Lorem ipsum, lorem ipsum,Lorem ipsum, lorem ipsum,Lorem ipsum}
    \end{subfigure}
    ~ 
      \makebox[\textwidth][c]{
      \begin{subfigure}[t]{1.4\linewidth}
    \centering
        \includegraphics[height=1.4in]{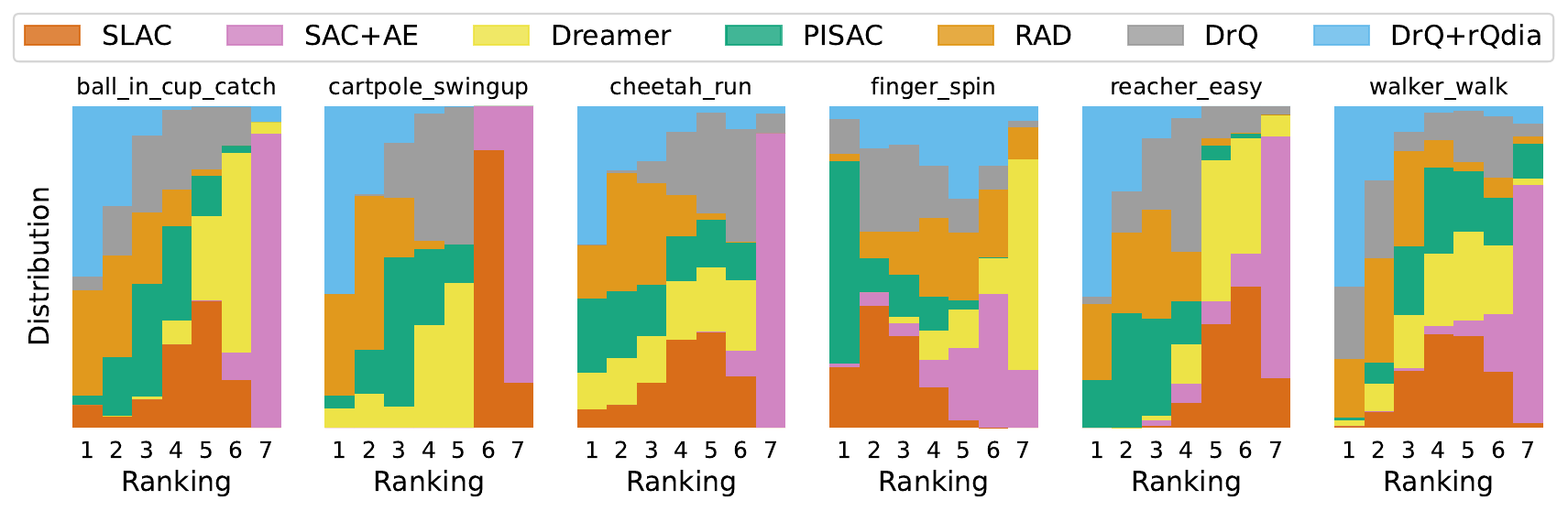}
        \caption{Lorem ipsum, lorem ipsum,Lorem ipsum, lorem ipsum,Lorem ipsum}
    \end{subfigure}
      }%
    
    \caption{Caption place holder}
\end{figure*}

\begin{figure*}[t!]
    \centering
    \begin{subfigure}[t]{0.5\textwidth}
        \centering
        \includegraphics[height=1.3in]{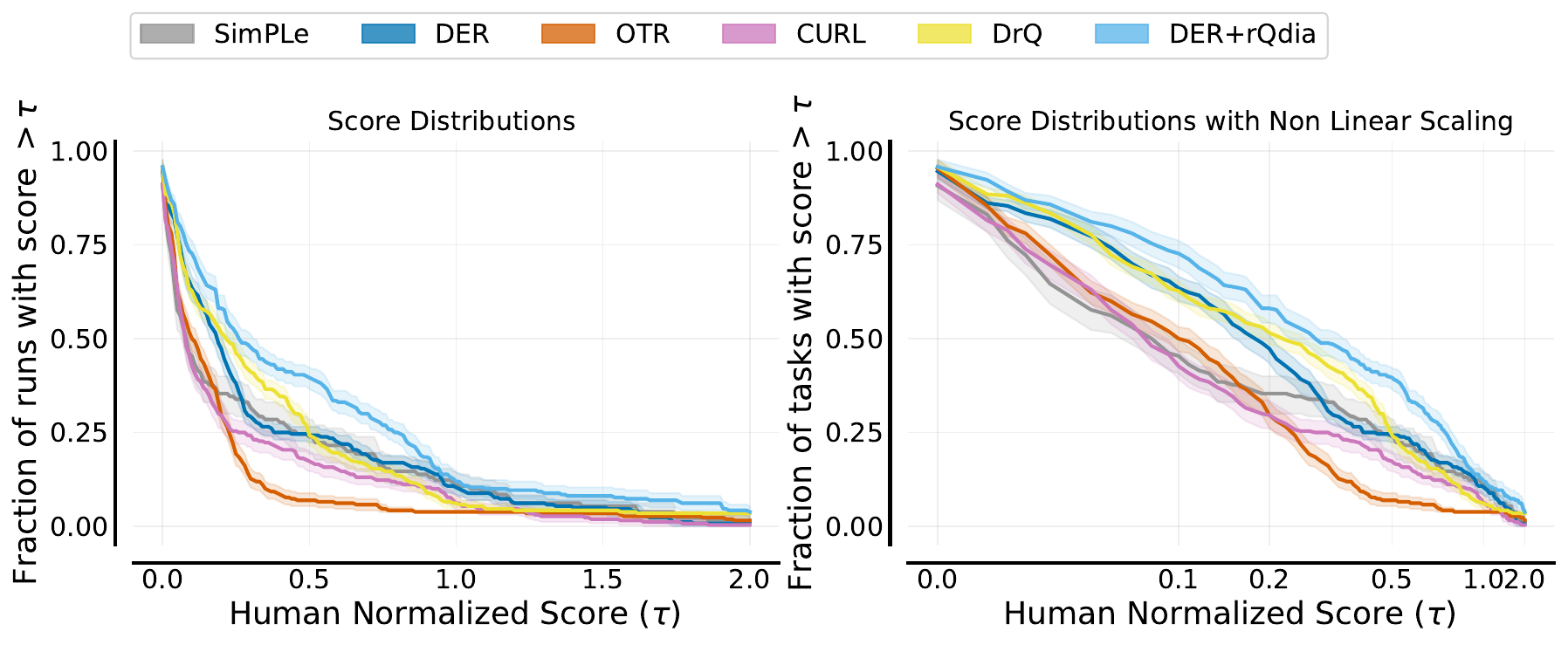}
        \caption{Lorem ipsum, lorem ipsum,Lorem ipsum, lorem ipsum,Lorem ipsum}
    \end{subfigure}
    ~
      \begin{subfigure}[t]{0.4\linewidth}
        \centering
        \includegraphics[height=1.4in]{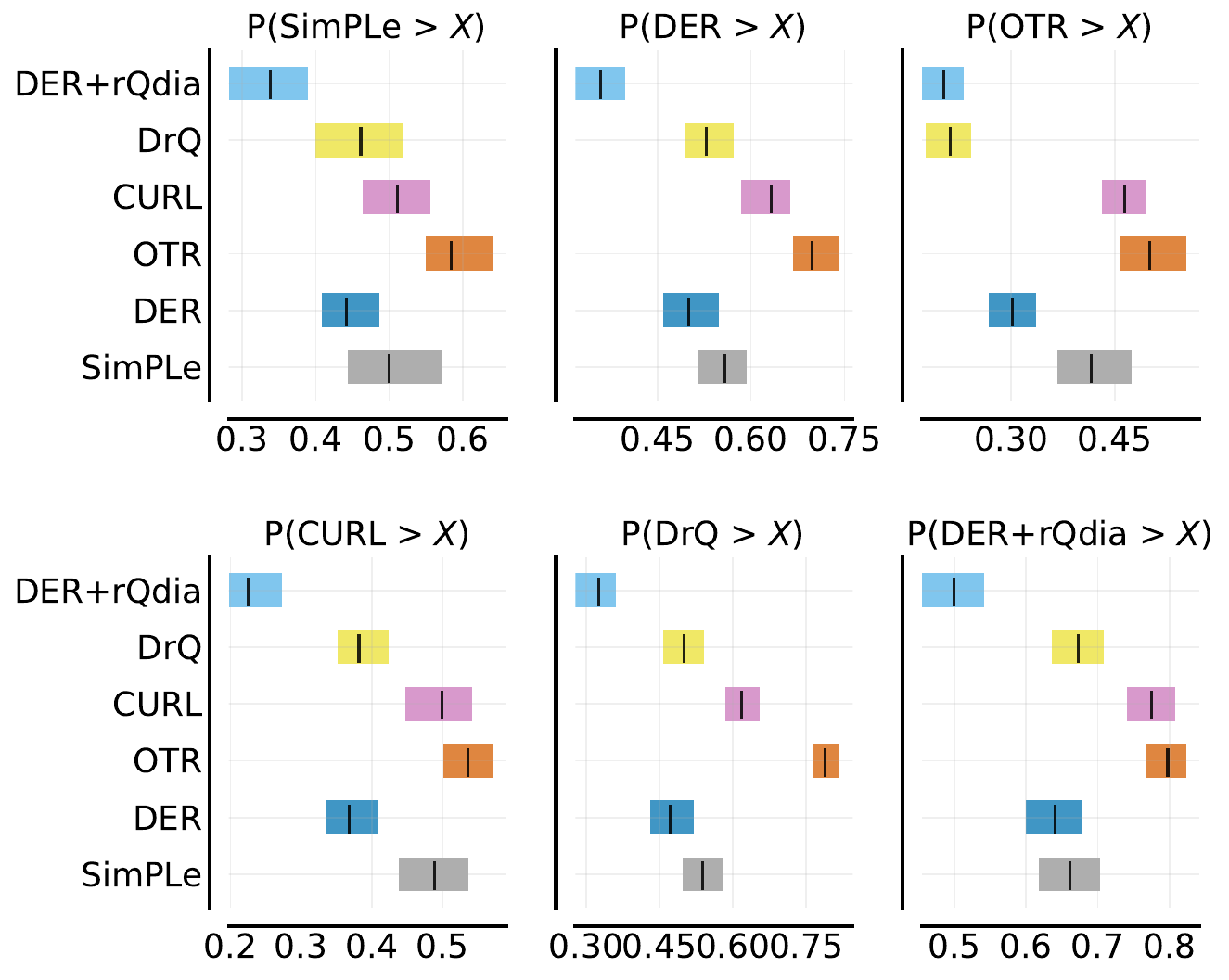}
        \caption{Lorem ipsum, lorem ipsum,Lorem ipsum, lorem ipsum,Lorem ipsum}
    \end{subfigure}
    ~
    ~
    ~
    \makebox[\textwidth][c]{
      \begin{subfigure}[b]{1\textwidth}
        \centering
        \includegraphics[height=1.1in]{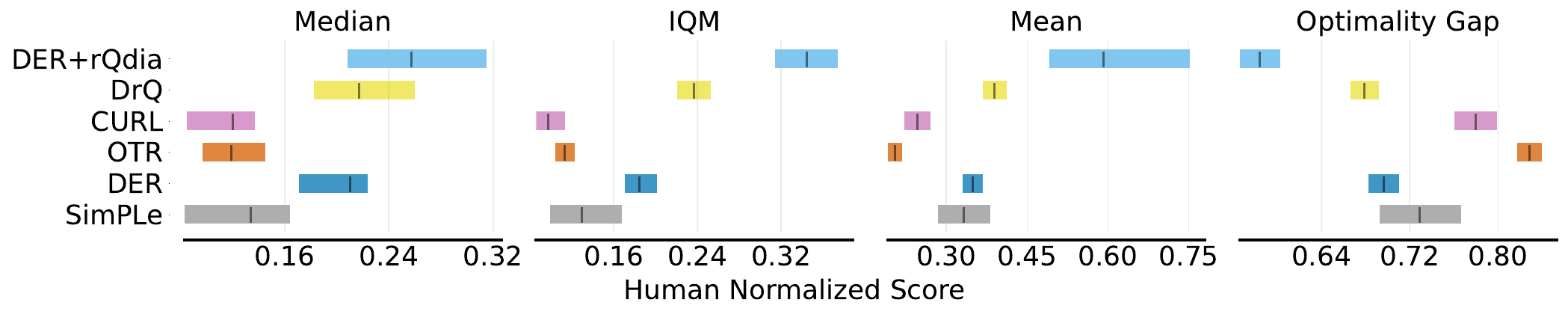}
        \caption{Lorem ipsum}
    \end{subfigure}%
    }%
    
    \caption{Caption place holder}
\end{figure*}

In the 500k continuous control setting, rQdia benefited training on 4/6 environments, as seen in Table~\ref{rqdiachart}. In those environments, we also saw performance roughly on par with or better than SAC from state embeddings on all 4/6 of these, compared to 1/6 for DrQ without rQdia. In both settings, rQdia combined with DrQ surpasses SAC from state embeddings more frequently and by wider margins generally than either DrQ~\citep{drq} or CURL~\citep{curl} standalone. In the Appendix, we provide full graphs of these runs.\\

\subsection{Atari Arcade 100k}




\begin{table}[t]
\caption{100k steps in the Atari Arcade Learning Environment. rQdia and CURL are both built on top of Data-Efficient Rainbow~\citep{effrainbow}. We report the best mean episode reward averaged across 3 random seeds, which is 2 seeds fewer than reported by DrQ~\citep{drq}.  We note that DrQ did not release their Atari code at time of submission.
rQdia improves upon Data-Efficient Rainbow on 18/26 environments, upon CURL on 15/26 environments, and achieves higher mean episode rewards and normalized scores. We reproduced CURL with the default hyperparams of the code released by~\citet{curl}, using the same hyperparams for rQdia+Eff. Rainbow as well. Scores for Eff. Rainbow, human, and random are copied over from those reported in DrQ~\citep{drq}. rQdia improves Eff. Rainbow by larger margins than CURL.}
\small
  \centering
  \begin{tabular}{lccccc}
\multicolumn{1}{c}{\bf Atari Arcade Environments}  &\multicolumn{1}{c}{\bf rQdia-Rainbow}
&\multicolumn{1}{c}{\bf CURL}
&\multicolumn{1}{c}{\bf Eff. Rainbow}
&\multicolumn{1}{c}{\bf Human}
&\multicolumn{1}{c}{\bf Random}
\\ \hline \\ 
        Alien&$\mathbf{1188}$ & 711.033 & 802.346 & 227.8 & 7127.7\\
Amidar&$\mathbf{208.9}$ & 113.743 & 125.905 & 5.8 & 1719.5\\
Assault&$\mathbf{649.9}$ & 500.927 & 561.46 & 222.4 & 742\\
Asterix&$\mathbf{890}$ & 567.24 & 535.44 & 210 & 8503.3\\
BankHeist&64 & 65.299 & $\mathbf{185.479}$ & 14.2 & 753.1\\
BattleZone&$\mathbf{19000}$ & 8997.8 & 8977 & 2360 & 37187.5\\
Boxing&$\mathbf{12.3}$ & 0.95 & -0.309 & 0.1 & 12.1\\
Breakout&8 & 2.555 & $\mathbf{9.214}$ & 1.7 & 30.5\\
ChopperCommand&$\mathbf{1500}$ & 783.53 & 925.87 & 811 & 7387.8\\
CrazyClimber&23970 & 9154.36 & $\mathbf{34508.57}$ & 10780.5 & 35829.4\\
DemonAttack&$\mathbf{1833}$ & 646.467 & 627.599 & 152.1 & 1971\\
Freeway&26.8 & $\mathbf{28.268}$ & 20.855 & 0 & 29.6\\
Frostbite&$\mathbf{2874}$ & 1226.494 & 870.975 & 65.2 & 4334.7\\
Gopher&$\mathbf{896}$ & 400.856 & 467.02 & 257.6 & 2412.5\\
Hero&$\mathbf{7261}$ & 4987.682 & 6226.044 & 1027 & 30826.4\\
Jamesbond&$\mathbf{985}$ & 331.05 & 275.66 & 29 & 302.8\\
Kangaroo&670 & $\mathbf{740.24}$ & 581.67 & 52 & 3035\\
Krull&$\mathbf{4193}$ & 3049.225 & 3256.886 & 1598 & 2665.5\\
KungFuMaster&$\mathbf{16310}$ & 8155.56 & 6580.07 & 258.5 & 22736.3\\
MsPacman&$\mathbf{1598}$ & 1064.012 & 1187.431 & 307.3 & 6951.6\\
Pong&-14.8 & -18.487 & $\mathbf{-9.711}$ & -20.7 & 14.6\\
PrivateEye&12.9 & $\mathbf{81.855}$ & 72.751 & 24.9 & 69571.3\\
Qbert&$\mathbf{2112.5}$ & 727.01 & 1773.54 & 163.9 & 13455\\
RoadRunner&8840 & 5006.11 & $\mathbf{11843.35}$ & 11.5 & 7845\\
Seaquest&$\mathbf{386}$ & 315.186 & 304.581 & 68.4 & 42054.7\\
UpNDown&$\mathbf{4154}$ & 2646.372 & 3075.004 & 533.4 & 11693.2\\
         \hline \\
         Better Mean Episode Reward & \textbf{16/26} & 10/26 & 2/26 & - & - \\
         Mean Human-Normalized Score &\textbf{17.36\%} &7.76\% &16.27\% & 100\% & 0\% \\
         Med Human-Normalized Score & \textbf{17.83\%}&13.37\% &13.99\% & 100\% & 0\% \\
  \end{tabular}
  \label{rqdiarainbow}
\end{table}

In the standard 100k efficient Atari benchmark, as reported by CURL~\citep{curl}, the addition of rQdia to Data-Efficient Rainbow~\citep{effrainbow} performs well compared to CURL, which relied on contrastive learning. We notice some remarkable improvements over both CURL and Eff. Rainbow on 15/26 environments, with an especially large gain for Frostbite, a game that depends heavily on translation invariance due to its visual symmetries. 

We note that since Rainbow's Q-value distribution is equivalent to its action distribution, that rQdia regularizes the action distribution directly here, while only doing so indirectly via Q-value distributions in continuous control. It is a question for future work whether regularizing actions directly in continuous control (equalizing their log probabilities) would benefit performance there.

We note that rQdia is orthogonal/complementary to CURL, DrQ, and other such methods, though we did not find performance improvements from rQdia in combination with CURL like we did with DrQ. DrQ has not yet released their Atari DQN code to the public.

\subsection{rQdia-SAC}



\begin{table}
\caption{Comparing rQdia added to just SAC-AE to state of the art models that succeeded SAC-AE. Similarly to Table \ref{rqdiachart}, 500k and 100k performances are shown, with our model disadvantaged in terms of batch size, and $3$ random seeds. To our surprise, results are largely on par with DrQ + rQdia, occasionally exceeding it, and other state of arts.}
  \centering
    \resizebox{\textwidth}{!}{%
  \begin{tabular}{lcccccccccc c}
    &
    \multicolumn{8}{c}{\textbf{From Pixels}} & \multicolumn{1}{c}{\textbf{State Emb}} \\
    \cmidrule(r){2-6}
    \cmidrule(r){7-7}
    \textbf{500k Step Scores}& DrQ+rQdia & DrQ & SLAC & SAC+AE & PISAC & RAD & Dreamer& CURL & PlaNet \\
\cmidrule(r){1-1}
Ball In Cup Catch&919.69 & 963.94 & 966.83 & 338.42 & 933.43 & $\mathbf{950.22}$ & 797.7 & 772 & 718\\
Cartpole Swingup&864.75 & 868.82 & 792.72 & 730.94 & 862.08 & 858.09 & $\mathbf{875.36}$ & 861 & 787\\
Cheetah Run&777.29 & 679.91 & 706.95 & 544.3 & $\mathbf{801.04}$ & 774.96 & 789.88 & 500 & 568\\
Finger Spin&939.05 & 938.77 & $\mathbf{983.34}$ & 914.3 & 967.56 & 907.4 & 793.26 & 874 & 718\\
Reacher Easy&$\mathbf{950.01}$ & 945.4 & 794.83 & 601.36 & 949.5 & 930.44 & 866.74 & 904 & 588\\
Walker Walk&934.47 & 924.16 & 951.58 & 858.16 & 934.21 & 917.58 & $\mathbf{956.22}$ & 906 & 478\\

    \hline  \\
    \textbf{100k Step Scores} &  &  &  &  &  &  \\

\cmidrule(r){1-1}
Ball In Cup Catch&910 & 913.8 & 917.56 & 338.42 & 933.43 & $\mathbf{950.22}$ & 797.7 & 772 & 718\\
Cartpole Swingup&$\mathbf{867.42}$ & 759.37 & 327.51 & 276.63 & 815.85 & 863.69 & 786.6 & 592 & 563\\
Cheetah Run&$\mathbf{502.77}$ & 360.97 & 413.49 & 240.58 & 459.86 & 499.06 & 422.54 & 307 & 165\\
Finger Spin&842.47 & 901.41 & 951.17 & 747.01 & $\mathbf{957.43}$ & 813.05 & 405.94 & 779 & 563\\
Reacher Easy&$\mathbf{905.34}$ & 600.42 & 342.5 & 225.14 & 757.85 & 772.44 & 373.08 & 517 & 82\\
Walker Walk&$\mathbf{721.78}$ & 633.57 & 528.96 & 395.87 & 513.9 & 644.78 & 533.06 & 344 & 221\\
  \end{tabular}
  }
 \label{rqdiasac}
\end{table}

We also evaluated how rQdia works independent of DrQ when augmented by itself as an auxiliary loss to SAC-AE. Indeed rQdia as a standalone auxiliary loss performs well, often even better than either DrQ and CURL. Please see Figure \ref{rqdiasac}.

While we note that rQdia is orthogonal to CURL, we found that combining the two did not benefit performance. Similarly, combining CURL with DrQ tended to hurt performance. rQdia and DrQ on the other hand, complemented each other well.

\section{Discussion}

\subsection{Limitations}

One limitation of rQdia is that it assumes a benefit to a certain augmentation invariance. Translation invariance for example might not be as useful in environments where most objects are held within a consistent axis. 

Moreover, rQdia ensures consistency between Q-value distributions across such perturbations, which means that models that do not need to learn such visual augmentations are expected to learn a more complex Q function.

On the other hand, such environments where these invariances are not useful or important may leverage rQdia to learn more invariant representations that could potentially better generalize to more complex environments.

\subsection{Ethics}

rQdia is a simple regularizer that contributes to the generalization of deep reinforcement learning models. While we hope deep RL continues to improve and its applications and abilities expand, we would be remiss not to note the destructive potential of the field, ranging from autonomous weapons to economic exploitation. However, we are optimistic that RL can do much more good than bad for society. Autonomous agents that can interact with the world and streamline robotics opens the door for countless medical, social, and economic benefits as well.

\subsection{Reproducability}

We have submitted Code together with the paper, a full pseudocode is offered in Algorithm \ref{rqdiacode} for reproducing continuous control, and a relevant snippet of Code is shared in Appendix \ref{acode} for reproducing Atari.


\section{Conclusion}

We presented a simple regularizer for model-free reinforcement learning that may easily be integrated into existing reinforcement learning frameworks. With the inclusion of this auxiliary loss, we attain strong performance compared to baseline models, including recent state of the arts. By regularizing Q-value distributions, we further enforce the invariances afforded by image augmentation techniques such that Q-value distributions are preserved under these perturbations. Consequently, we observe improvements in sample efficiency and final reward in the DeepMind Continuous Control Suite and environments in the Atari Arcade Learning Environments.


\bibliography{iclr2022_conference}
\bibliographystyle{iclr2022_conference}

\appendix
\section{Architecture}

The SAC-AE architecture we use is the same as [38], consisting of a shared encoder and distinct policy and Q-function heads. The CNN encoder is shared by the actor and critic; the critic consists of two ReLU-activated 3-layer MLP Q-networks; and the actor is a single ReLU-activated 3-layer MLP Gaussian policy head. We modify the code provided by [30]: \url{https://github.com/MishaLaskin/curl}.

Atari environments were tested with a Rainbow architecture inspired by [35] and built on the variant implemented in tandem with CURL in [30]. We added the rQdia auxiliary loss to their code sans the CURL-related portion. This code may be found here: \url{https://github.com/aravindsrinivas/curl_rainbow }.

\section{Training}

All hyperparameters were preserved from the original implementations discussed above. They are reviewed in Tables \ref{hyperparamsdrq} and \ref{hyperparamsrainbow}, except for the substitution of batch size since we used a batch size of $128$ while [30, 37] used $512$, giving those models a reasonable advantage.

\begin{table}[h]
\parbox{.45\linewidth}{
\centering
\begin{tabular}{lc}
    \toprule 
     Param & Value \\
	\midrule 
	Observation Size & (84, 84) \\
	Replay Buffer Size & 100000 \\
	Initial Steps & 1000 \\
	Stacked Frames & 3 \\
	Action Repeat & \makecell{2 finger, spin; \\walker, walk \\ 8 cartpole, swingup \\
4 otherwise} \\
MLP Hidden Units & 1024 \\
Evaluation Episodes & 10 \\
Optimizer & ADAM \\
Learning Rate $(f_\theta, \pi_\psi, Q_\phi)$ & 0.001 \\
Learning Rate $(\alpha)$ & 0.0001 \\
Batch Size & 128 \\
$Q$ Function EMA $\tau$ & 0.01 \\
Critic Target Update Freq & 2 \\
Conv Layers & 4 \\
Number of Filters & 32 \\
Non-Linearity & ReLU \\
Encoder EMA $\tau$ & 0.05 \\
Latent Dimension & 50 \\
Discount $\gamma$ & 0.99 \\
Initial Temperature & 0.1 \\
	\bottomrule
\end{tabular}
\caption{Hyperparameters for rQdia-DrQ}
\label{hyperparamsdrq}
}
\hfill
\parbox{.45\linewidth}{
\centering
\begin{tabular}{lc}
    \toprule 
     Param & Value \\
	\midrule 
	Observation Size & (84, 84) \\
	Replay Buffer Size & 100000 \\
	Frame Skip & 4 \\
	Action repeat & 4 \\
	Q-network Channels & 32, 64 \\
	Q-network Filter Size & $5 \times 5, 5 \times 5$ \\
	Q-network Stride & 5, 5 \\
	Q-network Hidden Units & 256 \\
	Momentum $\tau$ & 0.001 \\
	Non-Linearity & ReLU \\
	Reward Clipping & $[-1, 1]$\\
	Multi Step Return & 20 \\
	Min replay size\\ for sampling & 1600 \\
	Max Frames Per Episode & 108K\\
	Target network \\update period & 2000 updates\\
	Support Of Q-dist & 51 bins\\
	Discount $\gamma$ & 0.99\\
	Batch Size & 32\\
	Optimizer & ADAM \\
	Learning Rate & 0.9\\
	$(\beta_1, \beta_2)$ & (0.9, 0.999)\\
	Optimizer $\epsilon$ & 0.000015\\
	Max Grad Norm & 10\\
	Noisy Nets Parameter & 0.1\\
	Priority Exponent & 0.5\\
	Priority Correction & $0.4 \to 1$\\
	\bottomrule
\end{tabular}
\caption{Hyperparameters for rQdia-Rainbow}
\label{hyperparamsrainbow}
}
\end{table}

\section{Code}


\begin{figure}[h]
\centering
\includegraphics[width=.8\linewidth]{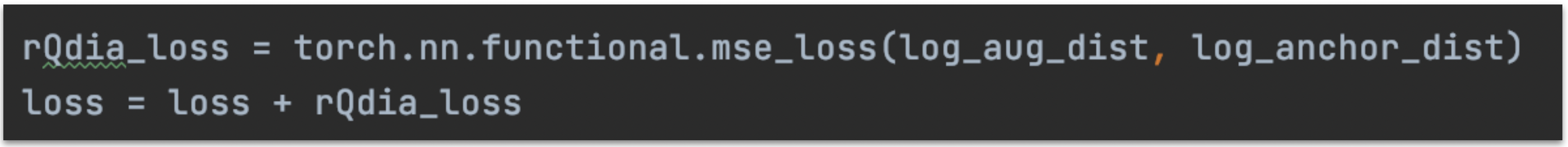}
\caption{Pytorch code for rQdia in Rainbow Atari.}
\label{rainbowrqdiacode}
\end{figure}

Code for continuous control and discrete Atari will be released on GitHub upon notification of decision and is provided separately together with the supplementary material.

The rQdia loss in Rainbow is a simple mean squared error between the anchor and augmentation's respective Q-value distributions, as shown in Figure \ref{rainbowrqdiacode}.

Continuous control is more complicated.



\section{Continuous Control}

\begin{figure}[h]
\centering
\includegraphics[width=\linewidth]{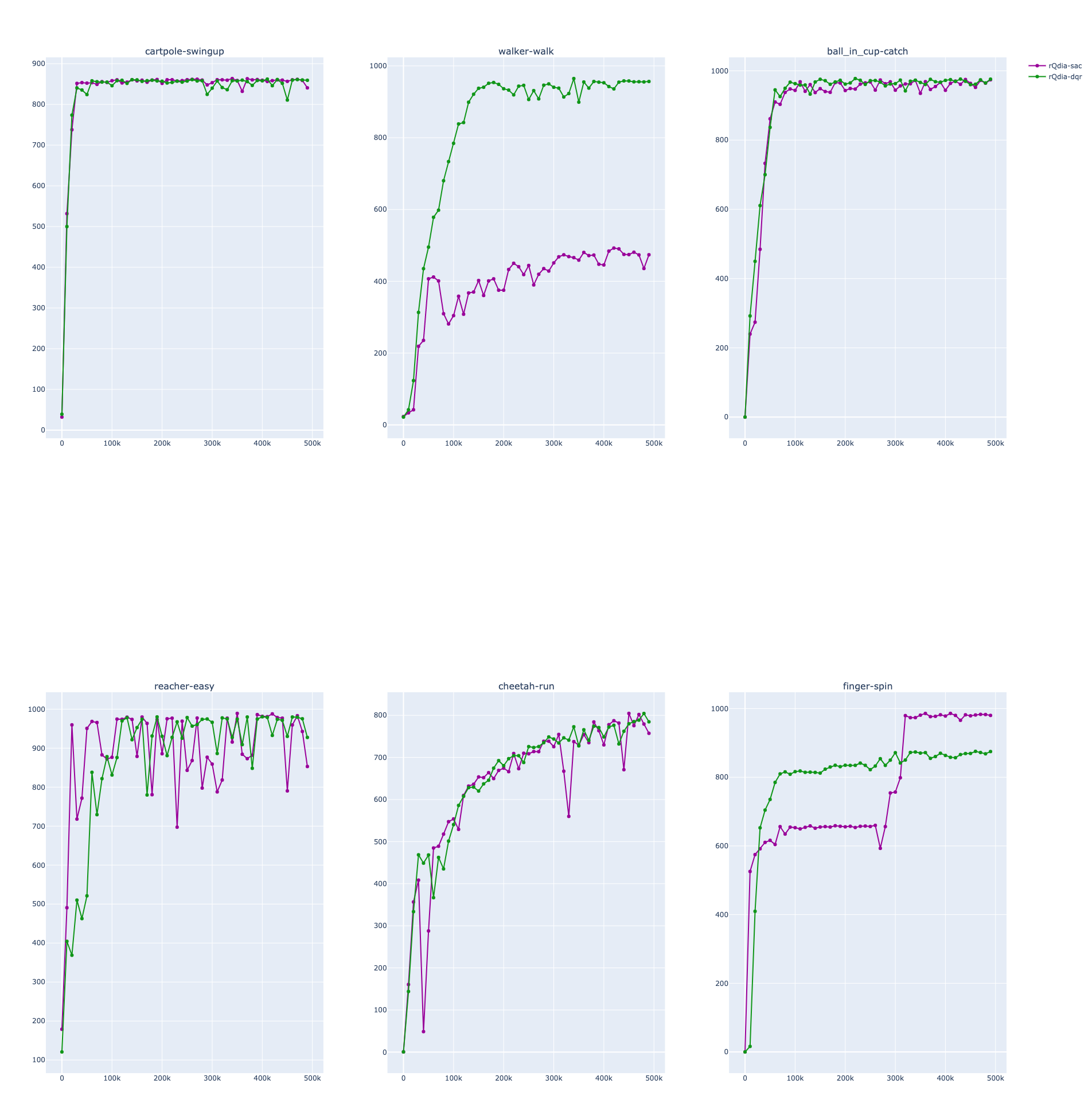}
\caption{Continuous control tasks.}
\label{mujoco}
\end{figure}

Plots of continuous control are shown in Figure \ref{mujoco}.

\section{Atari Arcade}

\begin{figure}[h]
\centering
\includegraphics[width=\linewidth]{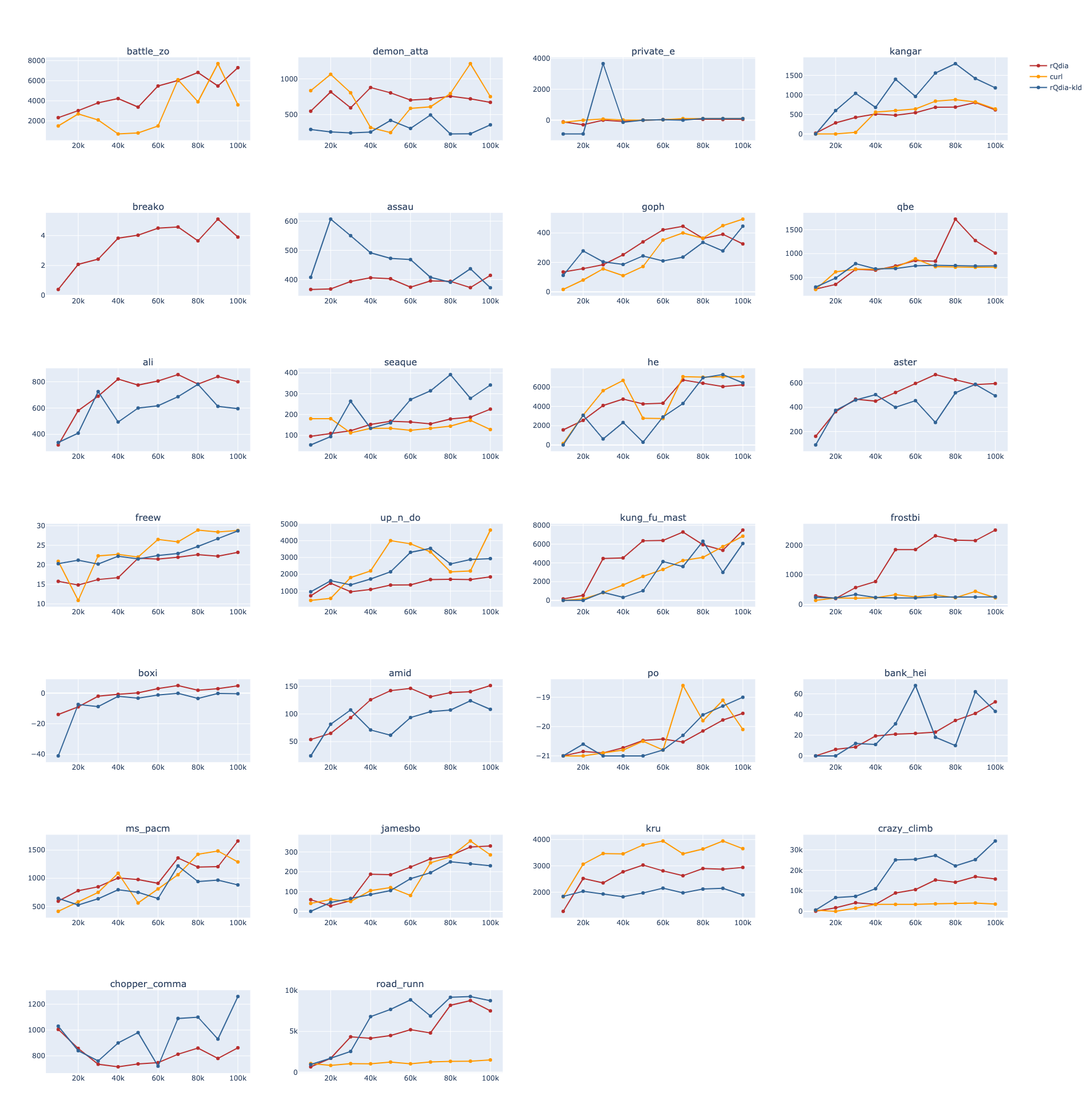}
\caption{Atari Arcade tasks.}
\label{rainbow}
\end{figure}

Plots for Atari Arcade are shown in Figure \ref{rainbow}.

\section{KL-Divergence Loss For rQdia-Rainbow}

\begin{table}[h]
\caption{KL divergence interestingly benefits mean human-normalized score, but is less optimal for other measures compared to mean squared error.}.
\small
  \centering
  \begin{tabular}{lccccc}
    \toprule 
     Atari Arcade Environments & rQdia-$D_{KL}$ & rQdia-$\mathit{MSE}$ & CURL-Rainbow & Human & Random  \\
	\midrule 

        Alien&$\mathbf{1188}$ & 711.033 & 802.346\\
Amidar&$\mathbf{208.9}$ & 113.743 & 125.905\\
Assault&$\mathbf{649.9}$ & 500.927 & 561.46\\
Asterix&$\mathbf{890}$ & 567.24 & 535.44\\
BankHeist&64 & 65.299 & $\mathbf{185.479}$\\
BattleZone&$\mathbf{19000}$ & 8997.8 & 8977\\
Boxing&$\mathbf{12.3}$ & 0.95 & -0.309\\
Breakout&8 & 2.555 & $\mathbf{9.214}$\\
ChopperCommand&$\mathbf{1500}$ & 783.53 & 925.87\\
CrazyClimber&23970 & 9154.36 & $\mathbf{34508.57}$\\
DemonAttack&$\mathbf{1833}$ & 646.467 & 627.599\\
Freeway&26.8 & $\mathbf{28.268}$ & 20.855\\
Frostbite&$\mathbf{2874}$ & 1226.494 & 870.975\\
Gopher&$\mathbf{896}$ & 400.856 & 467.02\\
Hero&$\mathbf{7261}$ & 4987.682 & 6226.044\\
Jamesbond&$\mathbf{985}$ & 331.05 & 275.66\\
Kangaroo&670 & $\mathbf{740.24}$ & 581.67\\
Krull&$\mathbf{4193}$ & 3049.225 & 3256.886\\
KungFuMaster&$\mathbf{16310}$ & 8155.56 & 6580.07\\
MsPacman&$\mathbf{1598}$ & 1064.012 & 1187.431\\
Pong&-14.8 & -18.487 & $\mathbf{-9.711}$\\
PrivateEye&12.9 & $\mathbf{81.855}$ & 72.751\\
Qbert&$\mathbf{2112.5}$ & 727.01 & 1773.54\\
RoadRunner&8840 & 5006.11 & $\mathbf{11843.35}$\\
Seaquest&$\mathbf{386}$ & 315.186 & 304.581\\
UpNDown&$\mathbf{4154}$ & 2646.372 & 3075.004\\
 Better Mean Episode Reward & 7/26 & \textbf{18/26} & /26 & - & - \\
 Mean Human-Normalized Score & \textbf{20.83\%} & 17.36\% &7.76\% & 100\% & 0\% \\
 Median Human-Normalized Score & 11.31\% & \textbf{17.83\%}&13.37\% & 100\% & 0\% \\
        
	\bottomrule
  \end{tabular}
  \label{rqdiarainbow}
\end{table}

We also evaluate rQdia as a KL divergence loss between the two distributional Q-value predictions yielded by Rainbow:

\begin{equation}
\label{rqdiaeqdkl}
\mathcal{L}_{rQdia} = D_{KL}(Q(s_t, a_i)||Q(aug(s_t), a_i)).
\end{equation}

While performance is not as good on all measures, as shown in Table \ref{rqdiarainbow}, KL divergence does benefit mean human-normalized score and yields significant improvements on particular environments; most notably, rQdia-$D_{KL}$ attains a score of $3655.80$ on PrivateEye, compared to $0$ for rQdia-$\mathit{MSE}$, 100 for CURL, and $69571.3$ for a human.

\end{document}